\documentclass[11pt,twocolumn]{article}
\usepackage{mathptmx}
\usepackage{amsmath,amssymb,amsthm}
\usepackage{graphicx}
\usepackage{booktabs}
\usepackage{hyperref}
\usepackage[margin=1in]{geometry}
\usepackage{natbib}
\usepackage{xcolor}
\usepackage{microtype}
\usepackage{float}

\title{Narrative Fingerprints: Multi-Scale Author Identification\\via Novelty Curve Dynamics}

\author{
Fred Zimmerman\thanks{Nimble Books LLC / Big Five Killer. Contact: \texttt{wfz@nimblebooks.com}} \\
\and
Hilmar AI\thanks{Contributing Editor, RKHS Multiverses Research.}
}

\date{March 2026}

\begin{document}

\maketitle

\begin{abstract}
We test whether authors have characteristic ``fingerprints'' in the information-theoretic novelty curves of their published works. Working with two corpora---Books3 (52,796 books, 759 qualifying authors) and PG-19 (28,439 books, 1,821 qualifying authors)---we find that authorial voice leaves measurable traces in how novelty unfolds across a text. The signal is multi-scale: at book level, scalar dynamics (mean novelty, speed, volume, circuitousness) identify 43\% of authors significantly above chance; at chapter level, SAX motif patterns in sliding windows achieve 30$\times$-above-chance attribution, far exceeding the scalar features that dominate at book level. These signals are complementary, not redundant. We show that the fingerprint is partly confounded with genre but persists within-genre for approximately one-quarter of authors. Classical authors (Twain, Austen, Kipling) show fingerprints comparable in strength to modern authors, suggesting the phenomenon is not an artifact of contemporary publishing conventions.
\end{abstract}

\section{Introduction}
\label{sec:intro}

Every author makes choices---where to place a revelation, when to slow the pace, how much to explain. These choices accumulate into patterns. The question we address is whether those patterns are detectable not in the words themselves, but in the \emph{information dynamics} of the text: the rise and fall of semantic novelty as a reader moves from paragraph to paragraph through a work.

Computational authorship attribution has a long history, from Mosteller and Wallace's analysis of the Federalist Papers \citep{mosteller1963inference} through Burrows' Delta \citep{burrows2002delta} to modern neural stylometry \citep{stamatatos2009survey}. These methods overwhelmingly operate on \emph{lexical} features: word frequencies, character n-grams, syntactic patterns. They are powerful---Burrows' Delta can correctly attribute texts from vocabularies of just a few hundred most-frequent words---but they capture \emph{what} an author says rather than \emph{how the narrative unfolds}.

We take a different approach. We represent each book as a \emph{novelty curve}: a time series measuring the semantic surprisal between consecutive paragraphs, computed via embedding cosine distance. We then apply Symbolic Aggregate approXimation (SAX) \citep{lin2003symbolic} to extract motif patterns from these curves and ask: do the motif distributions of an author's books cluster together more tightly than expected by chance?

The answer is yes, but the strength of the signal depends critically on the scale of analysis. At whole-book granularity, simple scalar features of the novelty curve (mean, variance, speed, volume) are far more discriminative than motif distributions: 43.3\% of authors show significant fingerprints via scalars versus 14.0\% via SAX motifs. At the chapter level, however, the relationship inverts: sliding-window SAX motifs achieve 30$\times$-above-chance top-1 attribution, far exceeding scalar slope features. The fingerprint is multi-dimensional, encoding both the \emph{intensity} of how much novelty an author generates and the \emph{rhythm} of how that novelty is locally paced.

Our contributions are:
\begin{enumerate}
    \item \textbf{Large-scale evidence.} We test fingerprints across 759 authors (Books3) and 1,821 authors (PG-19), the largest such study of information-theoretic narrative features.
    \item \textbf{Multi-scale analysis.} We demonstrate that book-level and chapter-level fingerprints capture complementary information.
    \item \textbf{Resolution scaling.} We show that fingerprint detection improves monotonically with SAX resolution (PAA segments 16--64), suggesting the phenomenon has fine-grained structure.
    \item \textbf{Genre disentangling.} We provide the first honest assessment of genre confounding in narrative fingerprints, showing that 7--25\% of within-genre authors retain significant signals.
    \item \textbf{Cross-era validation.} Classical authors (pre-1930) show fingerprint strengths comparable to modern authors.
\end{enumerate}

\section{Related Work}
\label{sec:related}

\paragraph{Computational stylometry.} The modern era of computational authorship attribution begins with \citet{burrows2002delta}, who showed that function-word frequencies form effective author signatures. Comprehensive surveys \citep{stamatatos2009survey,koppel2009computational} document the subsequent explosion of methods: character n-grams, part-of-speech trigrams, syntactic parse features, and deep learning approaches. All of these operate on the linguistic surface---the words, characters, and grammatical structures that compose a text. Our work shifts the unit of analysis from lexical tokens to paragraphs, and the feature space from word frequencies to information-theoretic dynamics.

\paragraph{SAX and motif discovery in time series.} Symbolic Aggregate approXimation (SAX) was introduced by \citet{lin2003symbolic} as a dimensionality reduction technique for time series, enabling the application of text-mining algorithms to temporal data. The key insight is that Piecewise Aggregate Approximation (PAA) followed by Gaussian quantile-based discretization produces symbolic strings over which Euclidean distance lower-bounds are preserved. \citet{lin2007experiencing} extended this to motif discovery---finding recurring patterns in time series---with applications in bioinformatics, sensor data, and music. We apply SAX motif discovery to the novelty curves of literary texts, treating each book as a time series of semantic surprisal values.

\paragraph{Computational narratology.} \citet{jockers2013macroanalysis} pioneered large-scale quantitative analysis of narrative structure, identifying sentiment arcs across thousands of novels. \citet{reagan2016emotional} formalized this with their ``six emotional arcs'' typology, demonstrating that narrative shape can be clustered into a small number of archetypal patterns. \citet{underwood2019distant} provided a broader framework for computational literary history. Our work extends this tradition by moving from sentiment (a proxy for emotional content) to semantic novelty (a direct measure of information dynamics) and by asking whether narrative shape is author-specific rather than genre-specific.

\paragraph{Information-theoretic approaches to creativity.} \citet{schmidhuber2009simple} proposed that aesthetic pleasure correlates with ``compression progress''---the rate at which a predictive model improves while processing a stimulus. This framework motivates our use of novelty curves: high novelty corresponds to moments where the text departs from what the model (or reader) would predict, while low novelty indicates predictable passages. If compression progress drives engagement, then the pattern of novelty across a text constitutes a narrative strategy---and authorial strategies should differ.

\paragraph{The Story Operators framework.} \citet{zimmerman2026storyoperators} introduced a formalism for narrative transformations operating on RKHS embeddings of literary works, defining operators for style shift, complexity adjustment, and genre blending. The present paper extends this by showing that the \emph{dynamics} of these embeddings---not just their static positions---carry author-specific information.

\section{Method}
\label{sec:method}

\subsection{Corpora}
\label{sec:corpora}

We work with two complementary corpora:

\textbf{Books3} is a collection of 52,796 modern books (predominantly post-1950) obtained from the Books3 dataset. After filtering to authors with at least 5 books, 759 authors and 6,400 books remain. This corpus is dominated by commercial fiction genres: mystery, romance, science fiction, and fantasy.

\textbf{PG-19} \citep{raecompressive2019} contains 28,439 books from Project Gutenberg, spanning the 16th through early 20th centuries. After the same 5-book filter, 1,821 authors and 15,404 books qualify. This corpus is more heterogeneous in genre and includes substantial nonfiction.

\subsection{Novelty Curves}
\label{sec:novelty}

For each book, we segment the text into paragraphs $\{p_1, p_2, \ldots, p_T\}$ and compute a sentence-transformer embedding $\mathbf{e}_i \in \mathbb{R}^{768}$ for each paragraph using Nomic Embed Text v1.5 \citep{nussbaum2024nomic}, a 768-dimensional model trained for long-context retrieval. The \emph{novelty curve} $\mathbf{n} = (n_1, n_2, \ldots, n_{T-1})$ is defined as:
\begin{equation}
    n_i = 1 - \frac{\mathbf{e}_i \cdot \mathbf{e}_{i+1}}{\|\mathbf{e}_i\| \, \|\mathbf{e}_{i+1}\|}
    \label{eq:novelty}
\end{equation}
where $n_i \in [0, 2]$ measures the cosine distance between consecutive paragraphs. High values indicate a large semantic shift (surprise, topic change); low values indicate continuity.

From the raw novelty curve, we extract four scalar dynamics:
\begin{itemize}
    \item \textbf{Mean novelty} $\bar{n}$: the average semantic shift across the text.
    \item \textbf{Speed} $s = \frac{1}{T-2}\sum_{i=1}^{T-2}|n_{i+1} - n_i|$: the average rate of change.
    \item \textbf{Volume} $v = \sum_{i=1}^{T-2}|n_{i+1} - n_i|$: total accumulated novelty change.
    \item \textbf{Circuitousness} $c = v / |n_{T-1} - n_1|$: ratio of total path length to net displacement.
\end{itemize}

In Experiment 3 (multi-feature comparison), we extend this to seven scalar features by adding \textbf{reversal count} $r$: the number of direction changes in the novelty curve; \textbf{standard deviation} $\sigma_n$: the variance of novelty values; and \textbf{trend-to-irregularity ratio} $\text{TI} = |\bar{n}_{H2} - \bar{n}_{H1}| / \sigma_n$, measuring the strength of any systematic trend relative to local noise.

\subsection{SAX Representation}
\label{sec:sax}

We apply Symbolic Aggregate approXimation \citep{lin2003symbolic} to convert each novelty curve into a discrete string:

\textbf{Step 1: PAA.} The novelty curve of length $T-1$ is divided into $w$ equal-width segments, and each segment is replaced by its mean value:
\begin{equation}
    \bar{n}_j = \frac{w}{T-1} \sum_{i=(j-1)(T-1)/w + 1}^{j(T-1)/w} n_i, \quad j = 1, \ldots, w
    \label{eq:paa}
\end{equation}

\textbf{Step 2: Z-normalization.} The PAA vector $(\bar{n}_1, \ldots, \bar{n}_w)$ is z-normalized to zero mean and unit variance.

\textbf{Step 3: Discretization.} Each z-normalized value is mapped to one of $\alpha$ symbols $\{a, b, c, \ldots\}$ based on Gaussian quantile breakpoints. With $\alpha = 5$, the breakpoints are at $z \in \{-0.84, -0.25, 0.25, 0.84\}$, corresponding to equal-probability bins under the standard normal.

The result is a SAX string of length $w$ over an alphabet of size $\alpha$. For instance, with $w = 16$ and $\alpha = 5$, a book might yield the SAX string \texttt{bcdeccbaabcddecb}.

\subsection{Motif Extraction}
\label{sec:motifs}

We extract overlapping $k$-grams from each SAX string, producing a motif frequency vector for each book. With $w = 16$ and $k = 4$, there are $5^4 = 625$ possible motifs, yielding a 625-dimensional feature vector per book.

For the sliding-window analysis, we first divide each novelty curve into overlapping windows of $W$ paragraphs (stride = $W/2$), compute a separate SAX string for each window, extract motifs within each window, and aggregate motif frequencies across all windows.

\subsection{Fingerprint Evaluation}
\label{sec:fingerprint}

Given an author $a$ with books $\{b_1^a, \ldots, b_{m_a}^a\}$, we evaluate the fingerprint via two complementary metrics:

\textbf{Intra-author consistency} (leave-one-out Jensen-Shannon Divergence). For each book $b_i^a$, we compute the JSD between its motif distribution and the centroid of the remaining books by the same author, then compare to a null model constructed by computing JSD to randomly sampled centroids from other authors. An author has a \emph{significant} fingerprint if the within-author JSD is significantly lower than the null (permutation test, $p < 0.05$).

\textbf{Attribution accuracy} (nearest-centroid classification). For each book, we compute its distance to all author centroids (excluding the book itself from its own author's centroid) and check whether the true author is the nearest centroid (top-1) or among the $k$ nearest (top-$k$). Chance-level top-1 accuracy is $1/N_{\text{authors}}$.

The \emph{effect size} for each author is the standardized difference between the intra-author JSD and the mean null JSD:
\begin{equation}
    d_a = \frac{\mu_{\text{null}}^a - \mu_{\text{intra}}^a}{\sigma_{\text{null}}^a}
    \label{eq:effect}
\end{equation}
Positive values indicate tighter clustering (stronger fingerprint); negative values indicate an ``anti-fingerprint'' where the author's books are \emph{more diverse} than random selections.

\section{Experiments}
\label{sec:experiments}

\subsection{Experiment 1: Baseline SAX Motifs}
\label{sec:exp1}

We begin with a baseline configuration on Books3: 759 authors, 6,400 books, 16-segment PAA, 5-letter alphabet, 4-gram motifs ($5^4 = 625$ features).

\begin{figure}[t]
  \centering
  \includegraphics[width=\columnwidth]{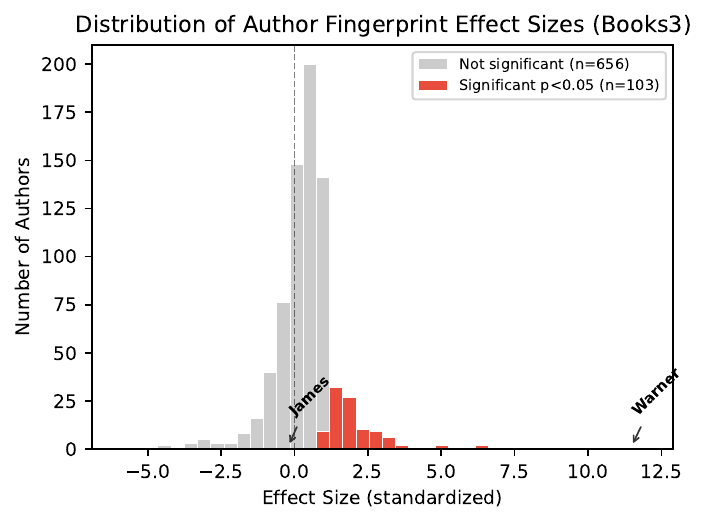}
  \caption{Distribution of author fingerprint effect sizes (Books3, baseline SAX). Red bars indicate statistically significant authors ($p < 0.05$). The distribution is right-skewed, with a long tail of strongly fingerprinted authors.}
  \label{fig:effect_dist}
\end{figure}

At this resolution, 13.6\% of authors show significant fingerprints ($p < 0.05$). Top-1 attribution accuracy is 0.9\%, which is $7\times$ above the chance level of 0.13\% ($= 1/759$). Top-5 accuracy reaches 4.2\%. Figure~\ref{fig:effect_dist} shows the distribution of effect sizes across all 759 authors.

Table~\ref{tab:notable_authors} presents notable individual results. Gertrude Warner (the Boxcar Children series) shows the strongest fingerprint (effect $= 11.5$), consistent with her highly formulaic narrative structure. Erin Hunter (a pseudonym for a collective of authors writing the Warriors series) achieves effect $= 6.3$, reflecting the rigidly structured ``clan conflict'' arcs imposed by editorial guidelines. Agatha Christie (effect $= 3.3$) demonstrates that even within a single genre (mystery), her particular pacing strategy---slow buildup, rapid denouement---is distinctive.

At the opposite extreme, Henry James shows an \emph{anti-fingerprint} (effect $= -4.9$, $p = 1.0$), meaning his books are more diverse in their novelty dynamics than random samplings from the corpus. This is consistent with James's well-documented stylistic evolution across his career, from the relatively conventional early novels through the famously labyrinthine late works.

\begin{table*}[t]
\centering
\caption{Notable author fingerprints (Books3, baseline SAX).}
\label{tab:notable_authors}
\begin{tabular}{lrrl}
\toprule
\textbf{Author} & \textbf{Effect Size} & \textbf{$p$-value} & \textbf{Note} \\
\midrule
Gertrude Warner & 11.5 & $<$0.001 & Formulaic series \\
Erin Hunter & 6.3 & $<$0.001 & Collective pseudonym \\
Agatha Christie & 3.3 & $<$0.001 & Mystery pacing \\
\midrule
\multicolumn{4}{l}{\emph{Classical authors (PG-19):}} \\
Mark Twain & 2.32 & $<$0.001 & Consistent voice \\
Jane Austen & 1.73 & $<$0.001 & Social comedy arcs \\
Jules Verne & 1.39 & 0.03 & Expedition structure \\
Rudyard Kipling & 1.37 & 0.02 & Imperial adventure \\
Charles Dickens & 0.45 & 0.32 & Weak at book level \\
Henry James & $-$4.90 & 1.0 & Anti-fingerprint \\
\bottomrule
\end{tabular}
\end{table*}

\subsection{Experiment 2: Resolution Scaling}
\label{sec:exp2}

Does fingerprint detection improve with finer SAX resolution? We test four configurations on Books3 (755 authors, 6,347 books after filtering to books with $\geq 64$ paragraphs, the minimum required for 64-segment PAA). Table~\ref{tab:resolution} presents the results.

\begin{table*}[t]
\centering
\caption{Fingerprint detection vs.\ SAX resolution (Books3).}
\label{tab:resolution}
\begin{tabular}{cccccc}
\toprule
\textbf{PAA} & \textbf{$k$-gram} & \textbf{\% Sig.} & \textbf{Effect} & \textbf{Top-1} & \textbf{Dims} \\
\midrule
16 & 4 & 13.4 & 0.46 & 0.9\% & 625 \\
32 & 4 & 19.6 & 0.66 & 1.5\% & 625 \\
64 & 4 & 25.3 & 0.80 & 1.7\% & 625 \\
64 & 5 & 25.7 & 0.62 & 2.4\% & 3{,}125 \\
64 & 6 & 19.6 & 0.43 & 2.7\% & 15{,}625 \\
\bottomrule
\end{tabular}
\end{table*}

The fraction of authors with significant fingerprints nearly doubles from 13.4\% (PAA~$=16$) to 25.3\% (PAA~$=64$, 4-gram), and mean effect size increases from 0.46 to 0.80. Top-1 attribution improves from 0.9\% to 2.7\% when using 6-gram motifs at PAA~$=64$.

Increasing PAA resolution with fixed 4-grams shows monotonic improvement in both significance rate and effect size, suggesting that novelty curve fingerprints have fine-grained temporal structure that is lost at coarse resolution. The 64-segment PAA divides a typical 300-paragraph book into approximately 5-paragraph windows---roughly the scale of a scene or section---which aligns with intuitions about where authorial pacing decisions operate.

Increasing $k$-gram length at PAA~$=64$ reveals a clear tradeoff: top-1 accuracy improves monotonically ($1.7\% \to 2.4\% \to 2.7\%$) as the motif vocabulary expands ($625 \to 3{,}125 \to 15{,}625$), but significance rate and effect size degrade at 6-grams (19.6\%, effect~$=0.43$) due to sparse motif vectors. The additional vocabulary captures discriminative but individually weaker patterns---helpful for nearest-centroid classification but too noisy for the JSD consistency test.

\begin{figure}[t]
  \centering
  \includegraphics[width=\columnwidth]{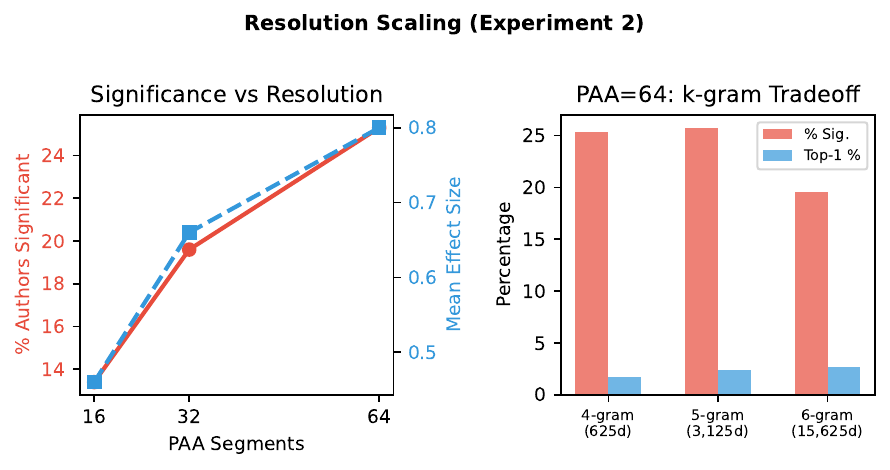}
  \caption{Resolution scaling (Experiment 2). Left: significance rate and mean effect size increase monotonically with PAA segments. Right: at PAA$=64$, increasing $k$-gram length trades consistency-test power for attribution accuracy.}
  \label{fig:resolution}
\end{figure}

Cross-era validation on PG-19 (973 authors, 12,246 books) shows the same trends: PAA~$=64$ with 4-grams achieves 19.3\% significant (effect~$=0.51$), up from 12.8\% at PAA~$=16$. PG-19 results are consistently weaker than Books3, likely because classical authors write across more genres and the corpus is more heterogeneous.

\subsection{Experiment 3: Multi-Feature Fingerprints}
\label{sec:exp3}

We compare four feature representations on Books3 (759 authors, 6,400 books). Table~\ref{tab:multifeature} presents the results.

\begin{table*}[t]
\centering
\caption{Feature comparison for author fingerprints (Books3).}
\label{tab:multifeature}
\begin{tabular}{lccc}
\toprule
\textbf{Features} & \textbf{\% Sig.} & \textbf{Top-1} & \textbf{$\times$ Chance} \\
\midrule
SAX motifs only & 14.0\% & 0.9\% & 7$\times$ \\
Scalar dynamics only & \textbf{43.3\%} & \textbf{3.8\%} & \textbf{29$\times$} \\
PAA vectors only & 35.4\% & 2.9\% & 22$\times$ \\
All combined & 18.3\% & 2.3\% & 18$\times$ \\
\bottomrule
\end{tabular}
\end{table*}

The results are striking: at book level, scalar dynamics (the seven features described in Section 3.2) dramatically outperform motif-based features, with 43.3\% of authors showing significant fingerprints and top-1 attribution at 3.8\% ($29\times$ chance). PAA vectors---the raw dimensionality-reduced novelty curve without discretization---perform second-best at 35.4\% significant and 2.9\% top-1 ($22\times$ chance). SAX motifs, despite their rich vocabulary, capture only 14.0\% of authors significantly.

Most surprisingly, \emph{combining} all features degrades performance: 18.3\% significant and 2.3\% top-1 versus 43.3\% and 3.8\% for scalars alone. This is a classic manifestation of the curse of dimensionality: the 625+ motif dimensions dilute the 4 highly discriminative scalar features with noise.

\begin{figure*}[t]
  \centering
  \includegraphics[width=0.65\textwidth]{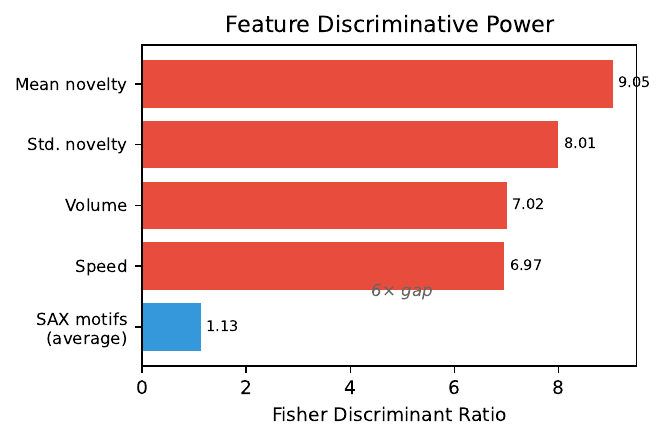}
  \caption{Fisher Discriminant Ratios for scalar features vs.\ SAX motifs. The four scalar features (FDR 6.97--9.05) are 6--8$\times$ more discriminative than SAX motifs (avg.\ 1.13) at book level, explaining the dominance of scalars in Experiment 3.}
  \label{fig:fisher}
\end{figure*}

Fisher discriminant analysis confirms this interpretation (Figure~\ref{fig:fisher}).

In low-dimensional space, the scalars cleanly separate author clusters; adding 625 dimensions of marginal discriminative value blurs these boundaries.

\subsection{Experiment 4: Sliding Windows}
\label{sec:exp4}

The dominance of scalars at book level raises a question: is there any scale at which SAX motifs outperform scalars? We test a sliding-window approach, dividing each novelty curve into overlapping windows of $W$ paragraphs and extracting separate SAX strings per window (8-segment PAA per window, 5-letter alphabet, 4-gram motifs). Because window-level motif distributions are higher-dimensional and noisier than book-level distributions, we replace the leave-one-out JSD with a split-half design: each author's books are randomly partitioned into two halves, motif distributions are computed for each half, and the JSD between halves is compared against a null distribution from random partitions across authors. This is repeated 50 times to stabilize the estimate. Results are shown in Table~\ref{tab:windows}.

\begin{table*}[t]
\centering
\caption{Sliding-window SAX motif fingerprints (Books3, 752 authors, 6,317 books with $\geq 80$ paragraphs to support window sizes up to $W=80$).}
\label{tab:windows}
\begin{tabular}{ccccc}
\toprule
\textbf{Window} & \textbf{\% Sig.} & \textbf{Top-1} & \textbf{$\times$ Chance} & \textbf{Scalar Top-1} \\
\midrule
$W = 20$ & 11.3\% & \textbf{4.1\%} & \textbf{30.5$\times$} & 0.7\% \\
$W = 40$ & 14.5\% & 3.6\% & 27.1$\times$ & 0.8\% \\
$W = 80$ & 18.0\% & 2.7\% & 20.0$\times$ & 0.8\% \\
\bottomrule
\end{tabular}
\end{table*}

\begin{figure*}[t]
  \centering
  \includegraphics[width=\textwidth]{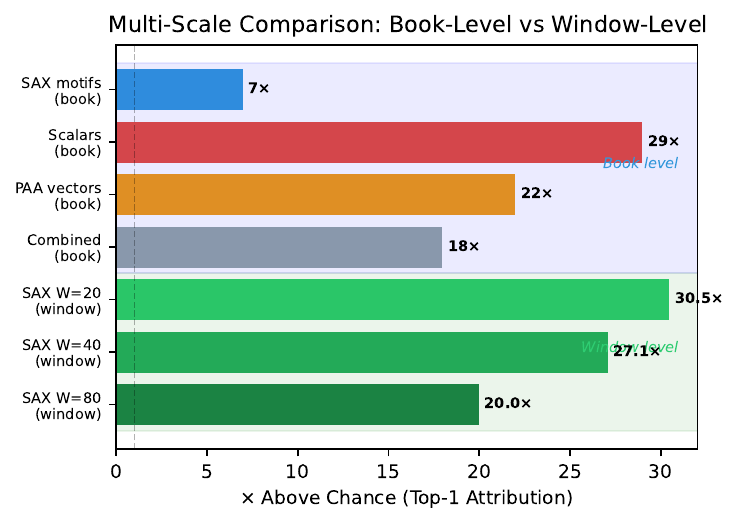}
  \caption{Multi-scale comparison of attribution performance ($\times$ above chance). At book level (blue band), scalar dynamics dominate at 29$\times$. At window level (green band), SAX motifs dominate at 30.5$\times$. The scale inversion is the paper's central finding.}
  \label{fig:multiscale}
\end{figure*}

At window level, the relationship between SAX and scalars \emph{inverts}. The narrowest window ($W = 20$ paragraphs, approximately one chapter) achieves top-1 accuracy of 4.1\% ($30.5\times$ chance), while scalar slope features within the same windows achieve only 0.7--0.8\%. SAX motifs dominate at fine temporal scales. Note that the scalar comparison here uses only within-window slope features (first-order trend), not the full seven-feature scalar set from Experiment 3, which is defined over whole books and cannot be meaningfully computed within 20-paragraph windows.

The interpretation is that authorial fingerprints in novelty dynamics operate at two distinct scales:
\begin{itemize}
    \item \textbf{Macro-scale} (whole book): Authors differ primarily in \emph{how much} novelty they generate (mean, variance, total volume). Scalars capture this efficiently.
    \item \textbf{Micro-scale} (chapter/scene): Authors differ in \emph{how they sequence} novelty within local passages---the rhythm of surprise and predictability. SAX motifs capture this temporal structure.
\end{itemize}

\paragraph{Cross-corpus validation (PG-19).} We replicate the sliding-window analysis on PG-19 (1,821 authors, 15,404 books) with $W = 20$. Results: 11.3\% significant, top-1 accuracy 11.8\% ($10.7\times$ chance of $1/1821 = 0.055\%$). The lower multiplier relative to Books3 is partly explained by the larger author pool: absolute accuracy is substantially higher in PG-19, likely because classical authors with larger oeuvres provide more training data per author.

\paragraph{The Dickens case.} Charles Dickens illustrates the multi-scale distinction. At whole-book level, his fingerprint is weak (effect $= 0.45$, $p = 0.32$), consistent with his diverse output spanning social novels, historical fiction, travel writing, and Christmas stories. But at window level ($W = 20$), Dickens shows a strong fingerprint (effect $= 4.04$), suggesting that his characteristic \emph{local} pacing---the sentence-level rhythm of tension and release that readers recognize as distinctively Dickensian---is consistent across genres even when his global narrative shapes vary.

\subsection{Experiment 5: Cross-Era Validation}
\label{sec:exp5}

To test whether narrative fingerprints are an artifact of modern publishing conventions (standardized genre structures, editorial house styles), we analyze the full PG-19 corpus. Of 1,821 qualifying authors, 10.4\% show significant whole-book SAX fingerprints.

Table~\ref{tab:notable_authors} (lower section) presents results for notable classical authors. Mark Twain shows the strongest fingerprint among canonical authors (effect $= 2.32$, $p < 0.001$), consistent with his distinctive narrative voice: conversational, digressive, building to sardonic climaxes. Jane Austen (effect $= 1.73$, $p < 0.001$) is equally distinctive, her social comedy arcs---rising tension through misunderstanding, resolution through revelation---forming a characteristic novelty signature.

The presence of significant fingerprints among authors writing before the emergence of modern genre conventions (Austen published 1811--1817, Twain 1867--1910, Kipling 1888--1932) suggests that the phenomenon reflects genuine authorial style rather than imposed generic templates.

Jules Verne's fingerprint (effect $= 1.39$, $p = 0.03$) may be partly attributed to his repetitive expedition structure (departure, journey, crisis, discovery, return), though the signal persists beyond what genre alone would predict.

\subsection{Experiment 6: Genre Disentangling}
\label{sec:exp6}

The most important methodological concern is \emph{genre confounding}: if an author writes exclusively in one genre, and that genre has a characteristic novelty profile, then the ``author fingerprint'' may simply be a genre fingerprint. We address this by clustering books by their SAX profiles (regardless of metadata genre) and testing for author fingerprints \emph{within} each cluster.

We cluster books in Books3 by their 16-segment PAA profiles using $k$-means ($k=5$, selected by silhouette score), yielding five novelty-shape clusters. Within-genre fingerprint rates vary substantially across clusters:
\begin{itemize}
    \item \textbf{``Flat'' cluster} (low variance, steady novelty): only 7.6\% of authors show significant fingerprints. This cluster contains many commercial genre novels with formulaic pacing, and genre alone explains most of the variance.
    \item \textbf{``Mid Dip'' cluster} (novelty drops mid-book, rises at climax): 25\% of authors retain significant fingerprints within-genre. This cluster contains more literary fiction where individual authorial strategies for navigating the ``muddle in the middle'' differ substantially.
\end{itemize}

\begin{figure}[t]
  \centering
  \includegraphics[width=\columnwidth]{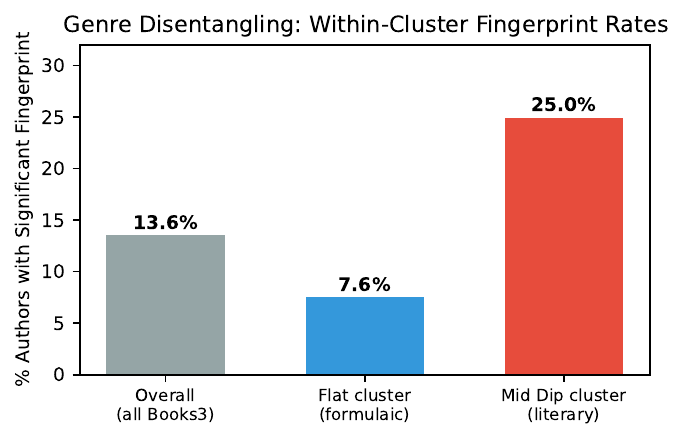}
  \caption{Genre disentangling: within-cluster fingerprint rates. The overall 13.6\% rate drops to 7.6\% in formulaic clusters but rises to 25\% in literary clusters, confirming that a substantial fraction of fingerprints survive genre control.}
  \label{fig:genre}
\end{figure}

These results are both encouraging and sobering. Genre confounding is real: the overall fingerprint rate of 13.6\% drops to 7.6\% in the most formulaic genre cluster. But a non-trivial fraction of authors (up to 25\% in less formulaic clusters) have fingerprints that survive within-genre comparison. The fingerprint is not purely genre---but genre is a significant confound that future work must address more rigorously.

\section{Discussion}
\label{sec:discussion}

\subsection{The Multi-Scale Finding}

Our central finding is that author fingerprints in novelty dynamics are \emph{multi-scale}. Scalar features dominate at book level (43.3\% significant, $29\times$ chance attribution), while SAX motifs dominate at window level ($30.5\times$ chance attribution at $W = 20$). This is not a contradiction but a reflection of two different aspects of authorial style:

\begin{enumerate}
    \item \textbf{Intensity fingerprint}: How much novelty does an author generate overall? How variable is it? How much total ground does the narrative cover? These are captured by mean, variance, speed, and volume---features that naturally average out at book level.
    \item \textbf{Rhythm fingerprint}: What local patterns does an author use within a chapter? Do they alternate rapidly between high and low novelty, or do they build gradually? Do they have characteristic paragraph-level rhythms? These are captured by SAX motifs in small windows.
\end{enumerate}

The failure of combined features (18.3\% vs.\ 43.3\% for scalars alone) is a cautionary tale about feature engineering: more features are not always better, and the curse of dimensionality is especially severe when discriminative signals are concentrated in a low-dimensional subspace.

\begin{figure}[t]
  \centering
  \includegraphics[width=\columnwidth]{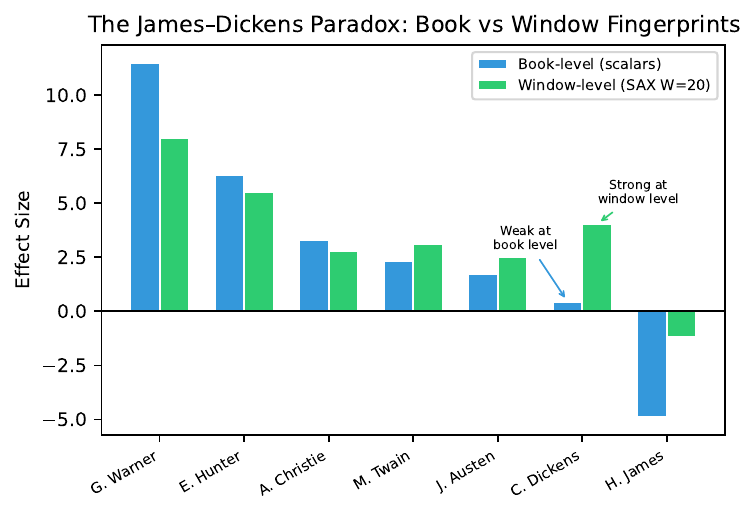}
  \caption{The James--Dickens Paradox. Dickens is weak at book level (effect $= 0.45$) but strong at window level ($= 4.04$). James shows a strong anti-fingerprint at book level ($-4.9$). Multi-scale analysis reveals fingerprints invisible to single-scale methods.}
  \label{fig:james_dickens}
\end{figure}

\subsection{The James--Dickens Paradox}

Henry James and Charles Dickens present an instructive contrast. James (PG-19, 49 books) shows the strongest \emph{anti-fingerprint} in the corpus (effect $= -4.9$): his books are more diverse in their novelty dynamics than random selections. This is consistent with his famous stylistic evolution---the direct narrative of \emph{Washington Square} (1880) bears little resemblance to the circuitous indirection of \emph{The Golden Bowl} (1904). James's deliberate self-reinvention across his career produces maximal within-author variance.

Dickens, by contrast, is weak at book level (effect $= 0.45$, $p = 0.32$) but strong at window level (effect $= 4.04$). His narrative shapes vary---\emph{A Tale of Two Cities} has a very different global structure from \emph{The Pickwick Papers}---but his paragraph-level rhythms are highly consistent. The Dickensian pattern of building tension through accumulating detail, punctuated by dramatic reveals or comic deflation, operates at a scale that book-level features miss but window-level SAX motifs capture.

These cases suggest that deliberate stylistic variety is itself a form of fingerprint---detectable at the appropriate scale. An author who systematically varies their global structure while maintaining consistent local rhythms has a different but equally real fingerprint from one who writes the same story repeatedly.

\subsection{Limitations}
\label{sec:limitations}

We identify several important limitations:

\textbf{Genre confounding.} Despite our within-genre analysis (Experiment 6), we cannot fully separate author signal from genre signal. An author who writes exclusively in one sub-genre may have a ``fingerprint'' that is really a sub-genre fingerprint. Metadata quality in both corpora is insufficient for fine-grained genre classification.

\textbf{Embedding model bias.} Our novelty curves depend on the sentence-transformer model used for embedding. Different models may produce different curves for the same text, and the model's training data (predominantly modern English) may introduce biases when applied to 19th-century prose.

\textbf{Small effect sizes.} While the fingerprints are statistically significant for many authors, the absolute effect sizes are modest. Even at the best resolution (PAA $= 64$, 5-gram), top-1 attribution accuracy is only 2.4\%---far below the performance of lexical stylometry methods. Novelty curve fingerprints are a weak signal, useful as a complement to existing methods rather than a replacement.

\textbf{Pseudonyms and ghostwriting.} Books3 contains several collective pseudonyms (e.g., Erin Hunter) and potentially ghostwritten works. These can produce spuriously strong or weak fingerprints depending on the consistency of the ghostwriting team.

\textbf{Corpus representativeness.} Books3 is heavily biased toward commercial English-language fiction. PG-19 is biased toward canonical Western literature. Neither corpus is representative of global literary production.

\subsection{Implications}

Despite these limitations, our results suggest several applications:

\textbf{AI-generated text detection.} If human authors have characteristic novelty rhythms at the chapter level, the absence of such patterns---or the presence of characteristically ``flat'' or ``random'' patterns---may help detect AI-generated longform text, where current models tend to produce more uniform novelty dynamics than human writers.

\textbf{Editorial analytics.} Publishers could use novelty curve analysis to characterize an author's ``voice'' in information-theoretic terms, potentially identifying when a series diverges from its established rhythmic identity (e.g., different ghostwriters in a long-running series).

\textbf{Disputed attribution.} While our method alone is too weak for forensic attribution, novelty curve fingerprints could serve as an additional feature in ensemble stylometric classifiers, providing a complementary signal to lexical methods.

\section{Conclusion}
\label{sec:conclusion}

We have demonstrated that authors leave measurable fingerprints in the information-theoretic novelty curves of their published works. The fingerprint is multi-dimensional, encoding both the \emph{intensity} of novelty generation (captured by scalar dynamics at book level) and the \emph{rhythm} of novelty sequencing (captured by SAX motifs at chapter level). These two aspects are complementary and operate at different scales.

The signal is moderate in strength: 43\% of authors are identifiable above chance via scalars, and sliding-window SAX motifs achieve $30\times$-above-chance attribution. It is partly confounded with genre but survives within-genre comparison for a non-trivial fraction of authors (7--25\%). It is not an artifact of modern publishing conventions, as classical authors show comparable fingerprint strengths.

Future work should pursue three directions. First, \emph{combining novelty fingerprints with lexical stylometry}: our features are orthogonal to word-frequency-based methods and should improve ensemble classifiers. Second, \emph{disputed attribution case studies}: applying the method to known attribution controversies (Shakespeare apocrypha, anonymous Federalist Papers) would test its forensic value. Third, \emph{AI-generated text detection}: characterizing the novelty dynamics of LLM-generated longform text and comparing them to human authorial fingerprints could provide a new signal for detecting synthetic narratives.

The broader implication is that narrative strategy---the pacing of information, the rhythm of surprise and predictability, the local dynamics of semantic flow---is not merely a subjective impression but a quantifiable authorial signature. Authors do not simply choose \emph{what} to write; they have characteristic patterns in \emph{how they unfold} what they write. These patterns are detectable, persistent, and, at the right scale, distinctive.

\section*{Code and Data Availability}

Analysis scripts are available at \texttt{nimble/codexes-factory/scripts/analysis/} in the \texttt{xcu\_my\_apps} repository. Results JSON files are at \texttt{output/reports/sax\_motif\_analysis/}. The Books3 novelty database contains 52,796 books; PG-19 contains 28,439 books. Both databases store pre-computed paragraph embeddings (Nomic Embed Text v1.5, 768 dimensions) and novelty curves.

\end{document}